\documentclass[11pt, a4paper]{article}
\usepackage[hyperref]{naaclhlt2019}
\usepackage{times}
\usepackage{latexsym}

\usepackage{url}  
\usepackage{graphicx}  

\usepackage{microtype}
\usepackage{subfigure}
\usepackage{booktabs} 
\usepackage{wrapfig}

\usepackage{hyperref}

\usepackage[utf8]{inputenc} 
\usepackage[T1]{fontenc}    
\usepackage{amsfonts}       
\usepackage{nicefrac}       
\usepackage{subfigure}
\usepackage{xcolor}
\usepackage{adjustbox}
\usepackage{bm}
\usepackage{amsmath}
\usepackage{amssymb}
\usepackage{algorithm}
\usepackage{algorithmic}
\usepackage{multirow}
\usepackage{adjustbox}

\newcommand{\RN}[1]{%
	\textup{\lowercase\expandafter{\it \romannumeral#1}}%
}

\aclfinalcopy

\usepackage{algorithm}
\usepackage{algorithmic}

\usepackage{amsmath}
\usepackage{amsfonts}
\usepackage{amssymb}

\newcommand{\beq}{\vspace{0mm}\begin{equation}}
\newcommand{\eeq}{\vspace{0mm}\end{equation}}
\newcommand{\beqs}{\vspace{0mm}\begin{eqnarray}}
\newcommand{\eeqs}{\vspace{0mm}\end{eqnarray}}
\newcommand{\barr}{\begin{array}}
\newcommand{\earr}{\end{array}}
\newcommand{\av}{{\boldsymbol a}}

\newcommand{\fv}{{\boldsymbol f}}

\newcommand{\sv}{{\boldsymbol s}}

\newcommand{\wv}{{\boldsymbol w}}

\usepackage[bottom]{footmisc}

\begin{document}
%
\title{Sequence Generation with Guider Network}

\author{Ruiyi Zhang$^1$, Changyou Chen$^2$, Zhe Gan$^3$, Wenlin Wang$^1$\\ 
	\textbf{Liqun Chen$^1$, Dinghan Shen$^1$, Guoyin Wang$^1$, Lawrence Carin$^1$}  \\
	$^1$Duke University, $^2$SUNY at Buffalo, $^3$Microsoft Research \\
	{\tt ryzhang@cs.duke.edu} \\}

\maketitle
\begin{abstract}
Sequence generation with reinforcement learning (RL) has received significant attention recently. However, a challenge with such methods is the sparse-reward problem in the RL training process, in which a scalar guiding signal is often only available after an entire sequence has been generated. This type of sparse reward tends to ignore the global structural information of a sequence, causing generation of sequences that are semantically inconsistent. In this paper, we present a model-based RL approach to overcome this issue. Specifically, we propose a novel guider network to model the sequence-generation environment, which can assist next-word prediction and provide intermediate rewards for generator optimization. Extensive experiments show that the proposed method leads to improved performance for both unconditional and conditional sequence-generation tasks.
\end{abstract}

\section{Introduction}
Sequence generation is an important area of investigation within machine learning. Recent work has shown excellent performance on a number of tasks, by combining reinforcement learning (RL) and generative models. Example applications include image captioning \cite{ren2017deep,rennie2016self}, text summarization \cite{li2018actor,paulus2017deep,rush2015neural}, and adversarial text generation \cite{guo2017long,lin2017adversarial,yu2017seqgan,zhang2017adversarial,zhu2018texygen}. The sequence-to-sequence framework (Seq2Seq) \cite{sutskever2014sequence} is a popular technique for sequence generation. However, models from such a setup are typically trained to predict the next token given previous ground-truth tokens as input, causing what is termed {\em exposure bias}  \cite{ranzato2015sequence}. By contrast, sequence-level training with RL provides an effective way to solve this challenge by treating sequence generation as an RL problem. By directly optimizing an evaluation score (cumulative rewards) \cite{ranzato2015sequence}, state-of-the-art results have been obtained in many sequence-generation tasks \cite{paulus2017deep,rennie2016self}. However, one problem in such a framework is that rewards in the RL training are particularly sparse, since a scalar reward is typically only available after an entire sequence has been generated. 

For RL-based sequence generation, most existing works rely on a model-free framework via recurrent policy gradient \cite{wierstra2010recurrent}. However,
these methods have been criticized for their high variance and poor sample efficiency \cite{sutton1998reinforcement}. On the other hand, model-based RL methods do not suffer from these issues, but they are usually difficult to train in complex environments. Furthermore, a learned policy is usually restricted by the capacity of an environment model. Recent developments on model-based RL \cite{kurutach2018model,nagabandi2017neural} combine the advantages of these two approaches, and have achieved improved performance by learning a model-free policy, assisted by an environment model. In addition, model-based RL has been employed recently to solve problems with extremely sparse rewards, with curiosity-driven methods \cite{pathak2017curiosity}.

Inspired by the ideas in \cite{kurutach2018model,nagabandi2017neural,pathak2017curiosity}, we propose a model-based RL method to overcome the sparse-reward problem in sequence-generation tasks. Our main idea is to employ a new guider network to model the generation environment in the feature space of sequence tokens, which is used to emit intermediate rewards by matching the predicted features from the guider network and features from generated sequences. The guider network is trained to encode global structural information of training sequences, useful to guide next-token generation in the generative process. Within the proposed framework, we also propose a new type of self-attention mechanism, to assist the guider network to provide high-level planning-ahead information. The intermediate rewards are combined with a final scalar reward, {\it e.g.}, an evaluation score in a Seq2Seq generation model or the discriminator loss in the generative-adversarial-net (GAN) framework, to train a sequence generator with policy-gradient methods. Extensive experiments show improved performance of our method on both unconditional and conditional sequence-generation tasks, relative to existing state-of-the-art methods.

\section{Background}
\subsection{Sequence-to-Sequence Model}
A sequence-generation model learns to generate a sequence $Y=(y_1, \ldots, y_T)$ conditioned on a possibly empty object $X$ from a different feature space. Here $y_t\in\mathcal{A}$ with $\mathcal{A}$ the alphabet set of output tokens. The pairs $(X,Y)$ are used for training a sequence-generation model. We use $T$ to denote the length of an output sequence, and $Y_{1,\dots,t}$ to indicate a subsequence of the form $(y_1, \ldots, y_t)$. The output of a trained generator is denoted $\hat{Y}$, which is typically not used during training. Since we focus on text generation in this paper, we will use {\em token} and {\em word} interchangeably to denote an element of a (text) sequence.

Starting from the initial hidden state $\sv_0$, a recurrent neural network (RNN) produces a sequence of states $(\sv_1,\ldots,\sv_T)$ given an input sequence-feature representation $(e(y_1), \dots, e(y_T))$, where $e(\cdot)$ denotes a function mapping a token to its feature representation. Let $e_t \triangleq e(y_t)$. The states are generated by applying a transition function $h: \sv_t=h(\sv_{t-1}, e_t)$ for $T$ times. The transition function $h$ is implemented by a cell of an RNN, with popular choices being the Long Short-Term Memory (LSTM) \cite{hochreiter1997long} and the Gated Recurrent Unit~(GRU) \cite{cho2014learning}. We will use the LSTM for our model. To generate a token $\hat{y}_t\in\mathcal{A}$, a stochastic output layer is applied on the current state $\sv_t$:  
\begin{align}
\hat{y}_t &\sim \text{Multi}(1, \text{softmax}(g(\sv_{t-1}))),\\
\sv_t &= h(\sv_{t - 1}, e(\hat{y}_t))\,,
\end{align}    
where $\text{Multi}(1,\cdot)$ denotes one draw from a multinomial distribution, and $g(\cdot)$ represents a linear transformation.
Since the generated sequence $Y$ is conditioned on $X$,  one can simply start with an initial state encoded from $X$: $\sv_0=\sv_0(X)$~\cite{bahdanau2016actor,cho2014learning}. 
Finally, a conditional RNN can be trained for sequence generation with gradient ascent by maximizing the log-likelihood of a generative model.

\subsection{Model-Based Reinforcement Learning}
Reinforcement learning is the problem of finding an optimal policy for an agent interacting with an unknown environment, collecting a reward per action. A policy is defined as a conditional distribution, $\pi(\av|\sv)$, defining the probability over an action $\av\in \mathcal{A}$ conditioned on a state variable $\sv\in\mathcal{S}$. Formally, the problem can be described as a Markov decision process (MDP), $\mathcal{M} = \langle\mathcal{S}, \mathcal{A}, P_s, r, \gamma\rangle$, where $P_s(\sv^\prime|\sv, \av)$ is the transition probability from state $\sv$ to $\sv^\prime$ given action $\av$; $r(\sv, \av)$ is an unknown reward function immediately following the action $\av$ performed at state $\sv$; $\gamma \in [0, 1]$ is a discount factor regularizing future rewards. At each time step $t$, conditioned on the current state $\sv_t$, the agent chooses an action $\av_t \sim \pi(\cdot|\sv_t)$ and receives a reward signal
$r(\sv_t, \av_t)$. The environment, as seen by the agent, then updates its state as $\sv_{t+1} \sim P_s(\cdot|\sv_t, \av_t)$. The goal is to learn an optimal policy such that one obtains the maximum expected total reward, {\it e.g.}, by maximizing the total rewards, 
\begin{align}\label{eq:policy_learning}
J(\pi) &= \sum_{t=1}^{\infty}\mathbb{E}_{P_s, \pi}\left[\gamma^t r(\sv_t, \av_t)\right]\\ 
&= \mathbb{E}_{\sv_t\sim\rho_{\pi}, \av_t\sim\pi}\left[r(\sv_t, \av_t)\right] \,,
\end{align}
where $\rho_{\pi} \triangleq \sum_{t=1}^{\infty}\gamma^{t-1}P_r(\sv = \sv_t)$, and $P_r(\sv)$ denotes the state marginal distribution induced by $\pi$.  

In model-based RL, a model of the
dynamics $P_s$ is built to make predictions for future states conditioned on the current state, which can be used for
action selection, {\it e.g.}, next-token generation. In practice, the model of dynamics is typically implemented as a discrete-time function, taking the current state-action pair $(\sv_t, \av_t)$ as input, and outputing an estimate of the next state $\sv_{t+\triangle t}$ at time $t+\triangle t$. At each step, the best next action is chosen based on the current policy, and the model will re-plan with the updated information from the dynamics. This control scheme is referred to as model-predictive control (MPC) \cite{nagabandi2017neural}. Note that in our setting, the state $\sv$ in RL typically corresponds to the current generated sentences $Y_{1,\ldots,t}$ in sequence generation, thus we use the same notation $\fv$ to denote both a state variable in RL and sequence generation. 


\section{Proposed Model}
\begin{figure}[t!] 
	\centering
	\includegraphics[width=\linewidth]{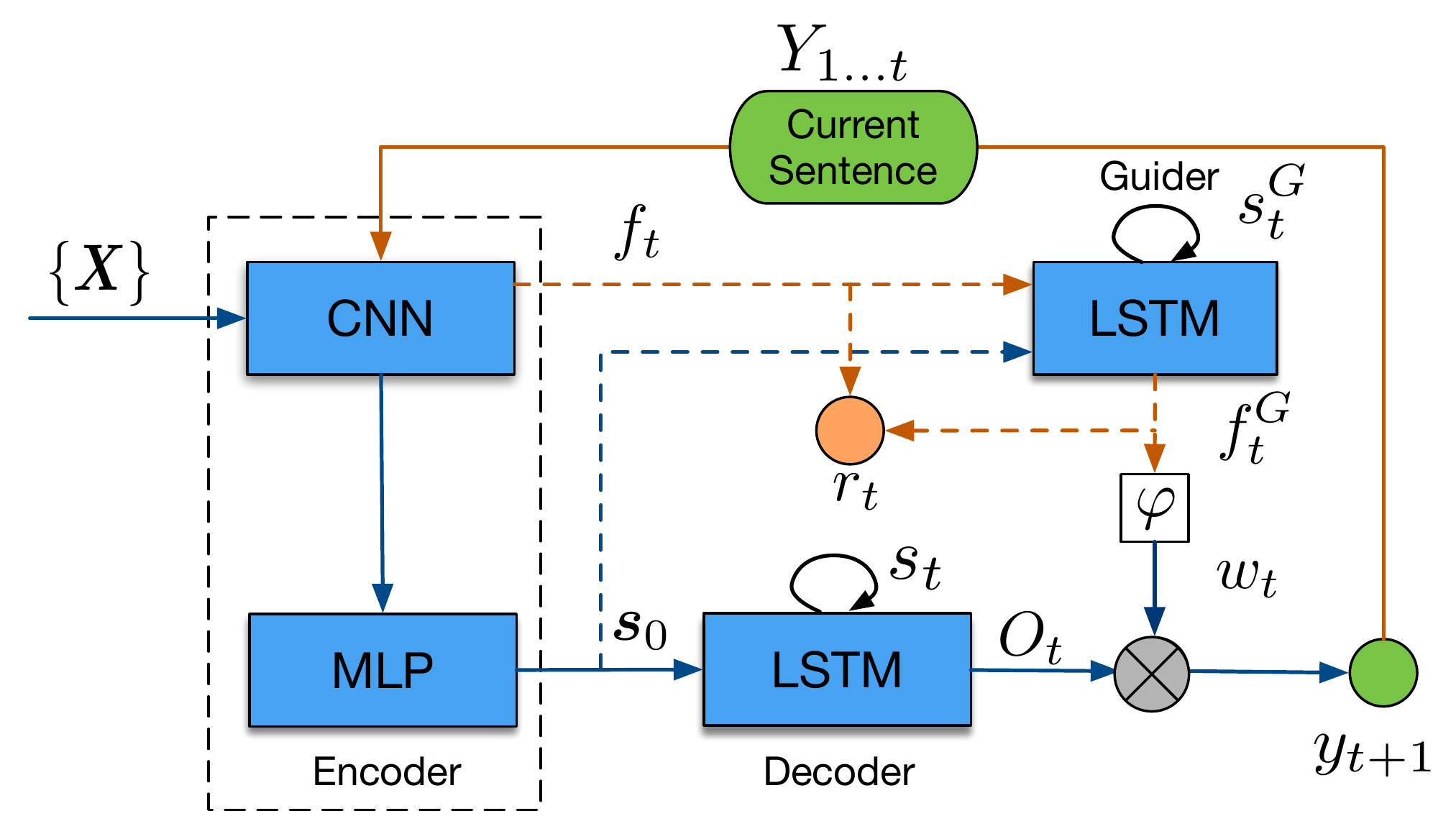}  
	\caption{Model Overview: Sequence generation with a guider network. Solid lines mean gradients are backpropagated in training; dash lines mean gradients are not backpropagated.}
	\label{fig:Model}
\end{figure}

To make the discussion explicit, we describe our model in the context of text generation, where tokens in a sequence are represented as words.
The model is illustrated in Figure~\ref{fig:Model}, with the first building block an Autoeocoder (AE) structure (the Encoder-Decoder in Figure~\ref{fig:Model}) for sentence feature extraction and generation. The encoder is shared for sentences from both training data and generated data, as explained in detail below. Overall, text generation can be formulated as a sequential decision-making problem.
At each timestep $t$, the agent, also called a generator (which corresponds to the LSTM Decoder in Figure~\ref{fig:Model}), takes the current LSTM state as the environment state, denoted as $\sv_t$. The policy $\pi^{\phi}(\cdot|\sv_t)$ parameterized by $\phi$ is a conditional generator, to generate the next token (action) given the state $\sv_t$ of current generated sequence. At each time step, an immediate reward $r_t$ is also revealed, which is calculated based on the output of the guider network and used to update the sequence generator described below. The objective of sequence generation is to maximize the total rewards, as in \eqref{eq:policy_learning}. We detail the components of our proposed model in the following subsections.
\subsection{Guider Network for Environment Dynamics}
The guider network, implemented as an RNN with LSTM units, is adopted to model environment dynamics to better assist sequence generation. The idea is to train a guider network such that its predicted sequence features at each time step are used to construct intermediate rewards in an RL setting, which in turn are used to optimize the sentence generator. Denote the guider network as $G^{\psi}(\sv_{t-1}^G, \fv_t)$, with parameters $\psi$ and input arguments $(\sv_{t-1}^G, \fv_t)$  at time $t$ to explicitly write out the dependency on the {\em guider network} latent state $\sv_{t-1}^G$ from the previous time step. Here $\fv_t$ is the input to the LSTM, which represents the feature of the current generated sentence after an encoder network. Specifically, let the current generated sentence be $Y_{1\ldots t}$ (forced to be the same as parts of a training sequence in training), with $\fv_t$ calculated as:
$\fv_t = \mbox{Enc}(Y_{1\ldots t})$,
where $\mbox{Enc}(\cdot)$ denotes the encoder transformation, implemented with a convolutional neural network (CNN)~\cite{zhang2017adversarial};
see Figure~\ref{fig:Model}. 
The initial state of the guider network is the encoded feature of a true input sequence by the same CNN, {\it i.e.}, $\sv_0^G = \mbox{Enc}(X)$.

\paragraph{Sequence-to-Sequence Generation with Planning}
We first explain how one uses the guider network to guide next-word generation for the generator (the LSTM decoder in Figure~\ref{fig:Model}). Our framework is inspired by the MPC method \cite{nagabandi2017neural}, and can be regarded as a type of plan-ahead attention mechanism. Given the feature $\fv_{t}$ at time $t$ from the current input sequence, the guider network produces a prediction $G^{\psi}(\sv^G_{t-1}, \fv_{t})$ as a future feature representation, by feeding $\fv_{t}$ into the LSTM. Since the training of the guider network is based on real data (detailed in the next paragraph), the predicted feature contains global-structure information of the training sequences. To utilize such information to predict the next word, we combine the predicted feature with the output of the decoder by constructing an attention-like mechanism. Specifically, we first apply a linear transformation $\varphi$ on the predicted feature $G^{\psi}(\sv^G_{t-1}, \fv_{t})$, forming a weight vector $\wv_t \triangleq \varphi\left(G^{\psi}(\sv^G_{t-1}, \fv_{t})\right)$. Next, the weight $\wv_t$ is applied to the output $O_t$ of the LSTM decoder by an element-wise multiplication operation. The result is then fed into a softmax layer to generate the next token $y_{t}$. Formally, the generative process is written as
\begin{align}
O_{t} &= g(\sv_{t-1}),\\
\wv_t &= \varphi(G^{\psi}(\sv^G_{t-1}, \fv_{t})),\\
y_{t}&\sim\text{Multi}(1,\text{softmax}(O_{t}\cdot\wv_t)),\\
\sv^G_{t} &= h^G(\sv^G_{t-1}, \fv_{t}),~~
\sv_{t} = h(\sv_{t-1}, e(y_{t}))~.
\end{align}
\paragraph{Guider Network Optimization} 
In training, given a sequence of feature representations $(\fv_1$, $\fv_2$, \ldots $\fv_T)$ for a training sentence, we seek to update the guider network such that it is able to predict $\fv_{t+c}$ given $\fv_{t}$, where $c > 0$ is the number of steps looked ahead. We implement this by forcing the predicted feature, $G^{\psi}(\sv_{t}^G, \fv_{t})$, to match both the sentence feature $\fv_{t+c}$ (first term in \eqref{eq:JG}) and the corresponding feature-changing direction (second term in \eqref{eq:JG}). This is formalized by maximizing an objective function of the following form at time $t$:
\begin{align}\label{eq:JG}
J^\psi_G= &\mathcal{D}_{\cos}\left(\fv_{t+c},~ G^{\psi}(\sv^G_{t-1}, \fv_{t})\right) +\\ &\mathcal{D}_{\cos}\left(\fv_{t+c}-\fv_{t},~ G^{\psi}(\sv_{t-1}^G, \fv_{t})-\fv_{t}\right)~,\nonumber
\vspace{-4mm}
\end{align}
where $\mathcal{D}_{\cos}(\cdot, \cdot)$ denotes the cosine similarity. By minimizing \eqref{eq:JG}, an ideal guider network should be able to predict the true next words conditioned on the current word in a sequence. As a result, the prediction is used to construct an intermediate reward, which is then used to update the generator (the LSTM decoder), as described further below. 
\subsection{Feature-Matching Rewards and Generator Optimization}
As in many RL-based sequence-generation methods, such as SeqGAN \cite{yu2017seqgan} and LeakGAN \cite{guo2017long}, the generator is updated based on policy-gradient methods. As a result, collecting rewards in the generation process is critical. Though \cite{yu2017seqgan} has proposed to use rollout to get rewards for each generated word, the variance of the rewards is typically too high to be practically useful. In addition, the computational cost may be too expensive for practical use.  We describe how to use the proposed guider network to define intermediate rewards below, leading to a definition of feature-matching reward.
\paragraph{Feature-matching rewards}
We first define an intermediate reward to generate a particular word. The idea is to match the ground-truth features from the CNN encoder in Figure~\ref{fig:Model} with those generated from the guider network. Equation \eqref{eq:JG} indicates that the further the generated feature is from the true feature, the smaller the reward should be. To this end, for each time $t$, we define the intermediate reward for generating the current word as:
\begin{align} 
r^g_t = \dfrac{1}{2c} \sum_{i=1}^{c} (&\mathcal{D}_{cos} (\fv_t, \hat{\fv}_t) +\\
 &\mathcal{D}_{cos} (\fv_t-\fv_{t-i}, \hat{\fv}_t-\fv_{t-i}))~, \nonumber
\end{align}
where $\hat{\fv}_t = G^{\psi}(s_{t-c-1}^G, \fv_{t-c})$ is the predicted feature.
Intuitively, $\fv_t-\fv_{t-i}$ measures the difference between the generated sequences in feature space; the reward will be high if it matches the predicted feature transition $\hat{\fv}_t-\fv_{t-i}$ from the guider network. At the last step of sequence generation, {\it i.e.}, $t = T$, the corresponding reward measures the quality of the whole generated sequence, thus it is called a final reward. The final reward is defined differently from the intermediate reward, which will be discussed below for both the unconditional- and conditional-generation cases.
%

Note a token generated at time $t$ will influence not only the rewards received at that time but also the rewards at subsequent time steps. Thus we propose to define the cumulative reward, $\sum_{i=t}^{T} \gamma^i r^g_i$ with $\gamma$ a discount factor, as a {\em feature-matching reward}. Intuitively, this encourages the generator to focus on achieving higher long-term rewards.
Finally, in order to apply policy gradient to update the generator, we combine the feature-matching reward with the problem-specific final reward, to form a $Q$-value reward specified below. We consider unconditional- and conditional-sequence generation in the following.
\paragraph{Unconditional generation} 
This case corresponds to generating sequences from scratch. Similar to SeqGAN, the final reward is defined as the output of a discriminator, evaluating the quality of the whole generated sequence, {\it i.e.}, the smaller the output, the less likely the generation is a true sequence. As a result, we combine the discriminator loss, denoted as $r^f$\footnote{This reward is given by the discriminator in GAN framework, and $r^f\in[0,1]$} with the feature-matching rewards as follows, to define a final $Q$-value reward:
\begin{align}
Q_t  = (\sum_{i=t}^{T} \gamma^i r^g_{i}) \times r^f ~.
\end{align}
\paragraph{Conditional generation}
This case corresponds to generating sequences conditioned on some input features, such as image features in image captioning. 
Following ideas from self-critical sequence training (SCST)~\cite{rennie2016self}, the final reward $r_s^f$ (also called the self-critical reward) is constructed by constituting a baseline reward, denoted as $\hat{r}^f(Y^\prime)$, for variance reduction: 
\begin{align}
r_s^f = r^f(Y) - \hat{r}^f(Y^\prime)~,
\end{align}  
where $r^f(Y)$ is the reward of a sampled sentence $Y$ by the current generator, and $ \hat{r}^f(Y')$ is the reward of a sentence obtained by choosing the words with the highest probabilities at each step $t$, {\it i.e.}, a greedy decoding.  Finally, the $Q$-value reward is defined as:
\begin{align}
Q_t  =\left\{  
\begin{array}{lr}  
r^f_s  \sum_{i=t}^{T} \gamma^i r^g_{i}  ~~~~~~~~~~~~~~~~~\text{if}~r_s^f > 0 \label{eq: conrewards}
\\
r^f_s \sum_{i=t}^{T} \gamma^i (1-r^g_{i}) ~~~~~~\text{otherwise}
\end{array}  
\right.  
\end{align}  
\paragraph{Generator optimization}
The sequence generator is initialized by pre-training on training sequences with an autoencoder structure, based on MLE training. After that, the final $Q$-value reward $Q_t$ is used as a reward for each time $t$, with standard policy gradient optimization methods to update the generator. Specifically, the policy gradient is
\begin{align}
\nabla_{\phi}J =	\mathbb{E}_{\sv_{t-1}\sim \rho_{\pi}, y_t \sim \pi}\left[ Q_t \nabla_{\phi} \log p(y_t| \sv_{t-1}; \phi_{t-1})\right] \,,
\end{align}
where $p(y_t| \sv_{t-1}; \phi_{t-1})$ is the probability of generating $y_t$ given $\sv_{t - 1}$ in the generator.
\paragraph{Discussion}
For unconditional generation, the feature-matching reward is typically good enough, since the task focuses more on sentence structure, which is reflected by the feature-matching reward. For conditional generation, however, a final reward in terms of an evaluation score ({\it e.g.}, the BLEU score) is more important because the semantic information of the conditioned variable is encoded into the score. This final reward thus guides the generator to generate semantically consistent sentence w.r.t.\! the conditioned variable. However, when most of the final rewards are negative, it is well-known that a policy gradient method would fail because of a lack of positive reward signals.

\subsection{Other Training Details}

\paragraph{Encoder as a feature extractor}
For unconditional generation, the feature extractor of the generating inputs for the guider network share the CNN part of the encoder. We stop gradients from the guider network to the encoder CNN in the training process. For conditional generation, we use a pre-trained feature extractor, trained similarly to the unconditional generation and fixed later on. 
\paragraph{Training procedure} 
As with many RL-based models~\cite{bahdanau2016actor,rennie2016self,sutskever2014sequence}, warm starting with a pre-trained model is important. Thus we first pre-train the encoder-decoder part based on the training data with an MLE loss. After pre-training, we use RL training to fine-tune the pre-trained generator. We adaptively transfer the training from MLE loss to RL loss, similar to~\cite{paulus2017deep,ranzato2015sequence}.
\paragraph{Initial states}
We use the same initial state for both the sequence generator and the guider network. For conditional generation, the initial state is the encoded latent code of the conditional information for both training and testing. For unconditional generation, it is the encoded latent code of a target sequence in training and random noise in testing.

\section{Related Work}

We first review related works that combine RL and GAN for text generation.
As one of the most representative models in this direction, 
SeqGAN~\cite{yu2017seqgan} adopts Monte-Carlo search to calculate rewards. However, such a method introduces high variance in policy optimization. There were a number of works proposed subsequently to improve the reward-generation process. For example, RankGAN~\cite{lin2017adversarial} proposes to replace the reward from the GAN discriminator with a ranking-based reward; MaliGAN \cite{cho2014learning} modifies the GAN objective and proposes techniques to reduce gradient variance; MaskGAN~\cite{fedus2018maskgan} uses a filling technique to define a $Q$-value reward for sentence completion; LeakGAN~\cite{guo2017long} tries to address the sparse-reward issue for long-text generation with hierarchical RL by utilizing the leaked information from a GAN discriminator. One problem of LeakGAN is that it tends to overfit on training data, yielding generated sentences that are often not diverse. By contrast, by relying on a model-based RL approach, our method learns global-structure information, which generates more-diverse sentences, and can be extended to conditional sequence generation.

RL techniques can also be used in other ways for sequence generation. For example, \cite{ranzato2015sequence} trains a Seq2Seq model by directly optimizing the BLEU/ROUGE scores with the REINFORCE algorithm. 
To reduce variance of the vanilla REINFORCE, \cite{bahdanau2016actor} adopts the actor-critic framework for sequence prediction. Furthermore, \cite{rennie2016self} trains a baseline with a greedy decoding scheme for the REINFORCE method. Note all these methods can only obtain rewards after a whole sentence is generated. Finally, planning techniques in RL have also been explored to improve sequence generation \cite{gulcehre2017plan,serdyuk2018twin}. Compared to these related works, the proposed guider network can provide a type of planning-ahead mechanism and intermediate rewards for RL training. Also, we consider using Q-value as the reward to encourage the generator focusing on long-term rewards.

\section{Experiments}
We test the proposed framework on unconditional and conditional sequence generation tasks, and analyze the results to understand the performance gained by the guider network.
%
%
We also perform an ablation investigation on the improvements brought by each part of our proposed method. All experiments are conducted on a single Tesla P100 GPU and implemented with TensorFlow. 

\subsection{Unconditional Text Generation}

We focus on adversarial text generation, and compare our approach with a number of related works \cite{guo2017long,lin2017adversarial,yu2017seqgan,zhang2017adversarial,zhu2018texygen}. In this setting, a discriminator in the GAN framework is added to the model in Figure~\ref{fig:Model} to guide the generator to generate high-quality sequences. This is implemented by defining the final reward to be the output of the discriminator. All baseline experiments are implemented on the texygen platform\footnote{https://github.com/geek-ai/Texygen}~\cite{zhu2018texygen}. We adopt the BLEU score, referenced by test set (test-BLEU) and themselves (self-BLEU)~\cite{zhu2018texygen} to evaluate quality of generated samples, where test-BLEU evaluates the reality of generated samples, and self-BLEU measures the diversity. A good generator should achieve both a high test-BLEU score and a low self-BLEU score. We call the proposed method feature-matching GAN (GMGAN) for unconditional text generation. A detailed description of GMGAN is provided in the Supplementary Material.
\paragraph{Short Text Generation: COCO Image Captions}
For this task, we use the COCO Image Captions Dataset \cite{lin2014microsoft}, in which most sentences are of length about 10. Since we consider unconditional text generation, only image captions are used as the training data. After preprocessing, the training dataset consists of 27,842 words and 417,126 sentences. We use 120,000 random sample sentences as the training set, and 10,000 as the test set. The BLEU scores with different methods are listed in Tables~\ref{tab:COCOun1} and \ref{tab:COCOunself}. We observe that
GMGAN performs significantly better than the baseline models. Specifically, besides achieving higher test-BLEU scores, the proposed method can also generate samples with very good diversity in terms of self-BLEU scores. LeakGAN represents the state-of-the-art in adversarial text generation, however, its diversity measurement is relatively poor \cite{zhu2018texygen}. We suspect the high BLEU score achieved by LeakGAN is due to its mode collapse on some good samples, resulting in high self-BLEU scores. Other baselines achieve lower self-BLEU scores since they cannot generate reasonable sentences.
\begin{table}[t!]
	\centering
	\begin{adjustbox}{scale=0.7,tabular=lcccc,center}
		\toprule[1.2pt]
		{Method}&BLEU-2&BLEU-3&BLEU-4&BLEU-5\\
		\midrule
		MLE & 0.820 & 0.607 & 0.389 & 0.248\\ 
		SeqGAN & 0.820 & 0.604 & 0.361 & 0.211\\
		RankGAN & 0.852 & 0.637 & 0.389 & 0.248\\
		GSGAN & 0.810 & 0.566 & 0.335 & 0.197\\
		LeakGAN & 0.922 & 0.797 & 0.602 & 0.416\\
		TextGAN & 0.926 & 0.774 & 0.552 & 0.362\\
		GMGAN (ours) & 0.949 & 0.823 & 0.635 & 0.421\\		
		\bottomrule[1.2pt]
	\end{adjustbox}
	\caption{\small 
		Test-BLEU scores on COCO Image Captions.
	}
	\label{tab:COCOun1}
\end{table}
\begin{table}[t!]
	\centering
	\begin{adjustbox}{scale=0.7,tabular=lccccccc,center}
		\toprule[1.2pt]
		{Method} & BLEU-2 & BLEU-3 & BLEU-4\\
		\midrule
		MLE & 0.754 & 0.511 & 0.232\\
		SeqGAN & 0.807 & 0.577 & 0.278\\
		RankGAN & 0.822 & 0.592 & 0.288\\
		GSGAN & 0.785 & 0.522 & 0.230\\
		LeakGAN & 0.912 & 0.825 & 0.689\\
		TextGAN & 0.830 & 0.597 & 0.284\\
		GMGAN (ours) & 0.746 & 0.511 & 0.319\\
		\bottomrule[1.2pt]
	\end{adjustbox}
	\caption{\small 
		Self-BLEU scores on COCO Image Captions.
	}
	\label{tab:COCOunself}
\end{table}

\begin{table*}[t!]
	\centering
	\begin{adjustbox}{scale=0.85,tabular= c p{6cm} p{11cm},center}
		\toprule
		\textbf{Method} & COCO Image Captions & EMNLP2017 WMT News\\ 
		\midrule
		\textbf{SeqGAN}& (1) A person made and black wooden table. \newline
		(2) A closeup of a window at night. & (1) She added on a page where it was made clear more old but public got said.\newline
		(2) I think she're guys in four years , and more after it played well enough. 
		\\ \hline
		\textbf{LeakGAN}&(1) A bathroom with a black sink and a white toilet next to a tub. \newline 
		(2) A man throws a Frisbee across the grass covered yard.&(1)"I'm a fan of all the game, I think if that's something that I've not," she said, adding that he would not be decided. \newline (2) The UK is Google' s largest non-US market, he has added "20, before the best team is amount of fewer than one or the closest home or two years ago.\\ \hline
		\textbf{GMGAN}& (1) Bicycles are parked near a row of large trees near a sidewalk. \newline
		(2) A married couple posing in front of a piece of birthday cake. & (1) "Sometimes decisions are big, but they're easy to make," he told The Sunday Times in the New Year.\newline (2) A BBC star has been questioned by police on suspicion of sexual assault against a 23-year-old man , it was reported last night.\\
		\bottomrule
	\end{adjustbox}
	\caption{\small 
		Examples of generated samples with different methods on COCO and EMNLP datasets.
	}
	\label{tab: WMTexample}
\end{table*}

\paragraph{Long Text Generation: EMNLP2017 WMT}
Following \cite{zhu2018texygen}, we use the News section in the EMNLP2017 WMT4 Dataset as our training data. 
The dataset consists of 646,459 words and 397,726 sentences. After preprocessing, the training
dataset contains 5,728 words and 278,686 sentences. The BLEU scores with different methods are provided in Tables \ref{tab:wmt} and \ref{tab:wmtself}. Compared with other methods, LeakGAN and GMGAN achieves comparable test-BLEU scores, demonstrating high quality of the generated sentences. Again, LeakGAN tends to over-fit on training data, leading to much higher (worse) self-BLEU scores. Our proposed GMGAN shows good diversity of long text generation with lower self-BLEU scores. Other baselines obtain both low self-BLEU and test-BLEU scores, leading to more random generations. 
\begin{table}[t!]
	\centering
	\begin{adjustbox}{scale=0.7,tabular=lccccccc,center}
		\toprule[1.2pt]
		{Method} & BLEU-2 & BLEU-3 & BLEU-4 & BLEU-5 \\
		\midrule  
		MLE & 0.761 & 0.468 & 0.230 & 0.116 \\
		SeqGAN & 0.630 & 0.354 & 0.164 & 0.087 \\
		RankGAN & 0.774 & 0.484 & 0.249 & 0.131 \\
		GSGAN & 0.723 & 0.440 & 0.210 & 0.107 \\
		LeakGAN & 0.920 & 0.725 & 0.502 & 0.321 \\
		TextGAN & 0.777 & 0.529 & 0.305 & 0.161 \\
		GMGAN (ours) & 0.923 & 0.727 & 0.491 & 0.303\\
		\bottomrule[1.2pt]
	\end{adjustbox}
	\caption{\small 
		Test-BLEU scores on EMNLP2017 WMT.
	}
	\label{tab:wmt}
\end{table}

\begin{table}[t!]
	\centering
	\vspace{2mm}
	\begin{adjustbox}{scale=0.7,tabular=lccccccc,center}
		\toprule[1.2pt]
		{Method} & BLEU-2 & BLEU-3 & BLEU-4 \\
		\midrule
		MLE & 0.664 & 0.338 & 0.113\\
		SeqGAN & 0.728 & 0.411 & 0.139 \\
		RankGAN & 0.672 & 0.346 & 0.119 \\
		GSGAN & 0.807 & 0.680 & 0.450 \\
		LeakGAN & 0.857 & 0.696 & 0.553 \\
		TextGAN & 0.806 & 0.662 & 0.448 \\
		GMGAN (ours) & 0.814 & 0.576 & 0.328 \\
		\bottomrule[1.2pt]
	\end{adjustbox}
	\caption{\small 
		Self-BLEU scores on EMNLP2017 WMT.
	}
	\label{tab:wmtself}
\end{table}
\paragraph{Human Evaluation}
Besides quantitatively evaluating the results using BLEU scores, we also conduct a human evaluation on WMT News dataset, using Amazon Mechanical Turk. 
%
In this regard, we randomly sample 100 sentences generated by each model. 10 human judges are asked to rate the generated texts in a scale from 0 to 5 in terms of their readability. The averaged human rating scores are shown in Table~\ref{tab:wmtscore}, indicating GMGAN achieves higher human scores compared to other methods. The averaged human rating scores are shown in Table~\ref{tab:wmtscore} and the results of GMGAN outperforms other methods. 
As examples, Table \ref{tab: WMTexample} illustrates some generated samples by GMGAN and its baselines. The performance on the two datasets indicates that the generated sentences of GMGAN are of higher global consistency and better readability than SeqGAN and LeakGAN. More generated examples are provided in the Supplementary Material.

\begin{table}[t!]
	\begin{adjustbox}{scale=0.80,tabular=c  c,center}
		\toprule[1.2pt]
		\textbf{Method} & \textbf{Human Score} \\
		\midrule
		MLE  & 2.45 \\
		SeqGAN & 2.57  \\
		RankGAN  & 2.91  \\
		GSGAN  & 2.48  \\
		LeakGAN  &  3.47  \\
		textGAN  & 3.11  \\
		\midrule
		GMGAN (ours) & \textbf{3.89} \\
		\bottomrule[1.2pt]
	\end{adjustbox}
	\caption{\footnotesize Human evaluation results on EMNLP2017 WMT.}
	\label{tab:wmtscore}
\end{table}

\begin{table*}[htp]
	\centering
	\small
	\begin{adjustbox}{scale=1,tabular=lcccccc,center}
		\toprule[1.2pt]
		\textbf{Method} & \textbf{BLEU-1} & \textbf{BLEU-2} & \textbf{BLEU-3} & \textbf{BLEU-4} & \textbf{METEOR} & \textbf{CIDEr} \\
		\midrule
		DeepVS~\cite{karpathy2015deep}  &    62.5 & 45.0 & 32.1 & 23.0 & 19.5 & 66.0 \\
		ATT-FCN~\cite{you2016image} &    70.9 & 53.7 & 40.2 & 30.4 & 24.3 & - \\
		Soft Attention~\cite{xu2015show}  & 70.7 & 49.2 & 34.4 & 24.3 & 23.9 & - \\
		Hard Attention~\cite{xu2015show}  &  71.8 & 50.4 & 35.7 & 25.0 & 23.0 & - \\
		Show \& Tell~\cite{vinyals2015show} & -  & -    & -    & 27.7 & 23.7 & 85.5 \\
		MSM~\cite{yao2016boosting} & 73.0 & 56.5 & 42.9 & 32.5 & 25.1 & 98.6 \\
		\midrule  
		\emph{No attention, Greedy, Tag} \\
		AE & 70.9 & 53.6 & 39.4 & 28.8 & 24.4 & 91.3 \\
		AE-g~ & 71.0 & 53.9 & 39.6 & 28.9 & 24.3 & 92.8 \\
		SCST (BLEU-4)~ & 73.2 & 57.1 & 43.9 & 33.6 & 24.5 & 95.9 \\
		GMST (BLEU-4)~ & 73.4 & 57.5 & \textbf{44.3} & \textbf{33.9}& 24.5 & 97.1 \\
		SCST (CIDEr)~ & 75.8 & 58.6 & 43.6 & 32.1 & 25.4 & 105.5 \\
		GMST (CIDEr)~ & \textbf{76.1} &\textbf{ 59.0} & 44.1 & 32.6 & \textbf{25.5} & \textbf{107.4} \\
		\midrule
		\emph{No attention, Greedy, Resnet-152} \\
		AE & 69.5 & 51.7 & 37.2 & 26.5 & 23.1 & 83.9 \\
		SCST (BLEU-4)~ & 71.1 & 54.8 & 41.6 & 31.6 & 23.1 & 87.5 \\
		GMST (BLEU-4)~ & 71.2 & 54.8 & \textbf{41.8}& \textbf{32.1} & 23.4 & 87.9 \\
		SCST (CIDEr)~ & \textbf{73.9} & 56.1 & 41.2 & 30.0 & 24.3 & 98.6 \\
		GMST (CIDEr)~ & 73.8 & \textbf{56.3} & 41.3 & 30.3 & \textbf{24.4} & \textbf{100.1} \\
		\bottomrule[1.2pt]
	\end{adjustbox}
	\caption{\small Results for image captioning on the MS COCO dataset; the higher the better for all metrics (BLEU 1 to 4, METEOR, and CIDEr). } 	\label{table:captioning}
\end{table*}
\subsection{Conditional Generation}
We conduct experiments on image captioning~\cite{karpathy2015deep} and text style transfer~\cite{shen2017style}. We investigate the benefits brought by the proposed method in (\ref{eq: conrewards}). In image captioning, instead of using a discriminator to define final rewards for generated sentence, we adopt evaluation metrics computed based on human references. 
The final rewards appear more important as they contain ground-truth information. Feature-matching rewards work as a regularizer to maintain the semantic consistency and sentence structure, preventing language-fluency damages caused by only focusing on evaluation metrics (final rewards). We call our model in this setting a feature-matching sequence training (GMST) model. An overview of GMST is provided in the Supplementary Material.

\paragraph{Image Captioning}
We first test our proposed model for image captioning on the MS COCO dataset \cite{lin2014microsoft}, which contains 123,287 images in total. Each image is annotated with at least 5 captions. Following Karpathy’s split \cite{karpathy2015deep}, 5,000 images are used for both validation and testing. We report BLEU-$k$ ($k$ from 1 to 4)~\cite{papineni2002bleu}, CIDEr~\cite{vedantam2015cider}, and METEOR~\cite{banerjee2005meteor} scores. We consider two settings: (\emph{i}) using a pre-trained 152-layer ResNet \cite{he2016deep} for feature extraction; 
(\emph{ii}) using semantic tags detected from the image as features~\cite{gan2017semantic}. We use an LSTM with 512 hidden units, and train the model with the Adam optimizer \cite{kingma2014adam}. The results are summarized in Table~\ref{table:captioning}. When comparing an  AutoEncoder (AE) with a variant implemented by adding a guider network (AE-$g$), reasonable improvements are observed. Next, we compare the proposed GMST with SCST, one of the state-of-the-art methods. Note the main difference between GMST and SCST is that the former employs our proposed feature-matching reward, while the latter only considers the final reward provided by evaluation metrics. GMST achieves higher scores compared with SCST on its optimized metrics.

\paragraph{Style Transfer}
We next test the proposed framework on the non-parallel text-style-transfer task, where the goal is to transfer one sentence in one style ({\it e.g.}, positive) to a similar sentence but with a different style ({\it e.g.}, negative). For a fair comparison, we use the same data and its split method as in~ \cite{shen2017style}. Specifically, there are 444,000, 63,500, and 127,000 sentences with either positive or negative sentiments in the training, validation and test sets, respectively. The task on this dataset is sentiment transfer.

\begin{table}[t!]
	\small
	\begin{adjustbox}{scale=0.95,tabular= lc,center}
		\toprule[1.2pt]
		\textbf{Method} & \textbf{Accuracy}\\
		\midrule
		VAE~\cite{shen2017style} & 23.2\%\\
		Cross-align~\cite{shen2017style} & 78.4\%\\
		CVAE~\cite{hu2017controllable} & 84.5\%\\
		\midrule
		Ours & \textbf{92.7\%}\\
		\bottomrule[1.2pt]
	\end{adjustbox}
	\caption{\small Sentiment accuracy of transferred sentences.}
	\label{tab: tranaccuracy}
\end{table}
%

\begin{table}[t!]
	\begin{adjustbox}{scale=0.6,tabular=l,center}
		\toprule[1.2pt]
		\textbf{From positive to negative}\\
		\midrule 
		Original: the food is amazing !  \\
		Transferred: the food is horrible ! 
		\vspace{2mm}
		\\
		Original: all the employees are friendly and helpful .  \\
		Transferred: all the employees are rude and unfriendly . \vspace{2mm}  
		\\
		Original: i 'm so lucky to have found this place ! \\
		Transferred: i 'm so embarrassed that i picked this place . \\
		\bottomrule[1.2pt]
		\textbf{From negative to positive}\\
		\midrule 
		Original: the service was slow . \\
		Transferred: the service was fast and friendly .\vspace{2mm}\\
		Original:  i would never eat there again and would probably not stay there either . \\
		Transferred:  i would definitely eat this place and i would recommend them . \vspace{2mm}\\
		Original: this place is aweful , everything about it was horrible !  \\ Transferred: this place is incredible , and incredibly good service .\\
		\bottomrule[1.2pt]
	\end{adjustbox}
	\caption{\small  
		Style transfer samples. The first line is the original sentence, the second is the generated sentences after style transfer.
	}
	\label{tab:transfer}
\end{table}

Table~\ref{tab:transfer} lists some sentiment-transfer examples. The proposed method can transfer sentiment while maintaining its original content.
To measure whether the original sentences (in the test set) have been transferred to the desired sentiment, we follow the settings of~\cite{shen2017style} and employ a pretrained CNN classifier, which achieves an accuracy of $97.4\%$ on the validation set, to evaluate the transferred sentences. 
Results are shown in Table~\ref{tab: tranaccuracy}. 
It can be observed that our proposed model exhibits much higher transfer accuracy, indicating the guider network provides good
 sentiment guidance for the generator.
%




\section{Conclusion}
We propose an RL-based method for learning a sequence model, by introducing a guider network to model the generation environment.
The guider network provides a plan-ahead mechanism for next-word selection. Furthermore, feature rewards are calculated based on the guider network, to overcome the sparse-reward problem in previous methods; they are used to optimize the generator via policy-gradient method. Our proposed models are validated on both unconditional and conditional sequence generation, including adversarial text generation, image captioning and style transfer. We obtain state-of-the-art results in terms of generation quality and diversity for unconditional generation, and achieve improved performance on several conditional-generation tasks.


\bibliography{reference}
\bibliographystyle{aaai}
\newpage
\appendix 

\title{Supplementary Material \\Sequence Generation with a Guider Network} 
\maketitle
\newpage .
\newpage

\section{Extensive Experiments}
\subsection{Ablation Study}
We conduct ablation studies on long text generation to investigate the improvements brought by each part of our proposed method. First, we test the benefits of using the guider-network information for word selection. Among the methods compared, VAE-g is the standard VAE with the guider network as a pre-trained baseline. We compare RL training with $\RN{1})$ only final rewards\footnote{We only use RL training for 200 batches, as the performance keeps dropping with more training time.}, $\RN{2})$ only feature-matching rewards, and $\RN{3})$ combining both rewards, namely GMGAN. The results are shown in Table~\ref{tab:ablation} and Table~\ref{tab:ablationself}. We observe that guider network plays an important role on improving the baselines. RL training with final rewards given by a discriminator typically damages the generation quality; whereas feature-matching reward produces sentences with much better diversity due to the ability of exploration. 
\begin{table}[H]
	\centering
	\begin{adjustbox}{scale=0.7,tabular=l c  c  c  c c  , center}
		\toprule[1.2pt]
		{Method} & MLE & VAE-g & Final & Feature &  GMGAN\\
		\midrule
		BLEU-2  &0.761& 0.920 & 0.843 &  0.914  & 0.923\\
		BLEU-3  &0.468& 0.723 & 0.623 &  0.704  & 0.727\\
		BLEU-4  &0.230& 0.489 & 0.390 &  0.457  & 0.491\\
		BLEU-5  &0.116& 0.289 & 0.221 &  0.276  & 0.303\\
		\bottomrule[1.2pt]
	\end{adjustbox}
	\caption{\small
		BLEU scores on EMNLP2017 WMT.
	}
	\label{tab:ablation}
\end{table}

\begin{table}[H]
	\centering
	\begin{adjustbox}{scale=0.7,tabular=l  c  c  c  c  c  , center}
		\toprule[1.2pt]
		{Method} &MLE& VAE-g & Final & Feature &  GMGAN\\
		\midrule
		BLEU-2  &0.664& 0.812 & 0.778 &  0.798  & 0.814\\
		BLEU-3 &0.338& 0.589 & 0.525 &  0.563  & 0.576\\
		BLEU-4  &0.113& 0.360 & 0.273 &  0.331  & 0.328\\
		\bottomrule[1.2pt]
	\end{adjustbox}
	\caption{\small
		Self-BLEU scores on EMNLP2017 WMT.
	}
	\label{tab:ablationself}
	
\end{table}
\subsection{Illustrations of Feature Matching Rewards}
\begin{figure}[ht] \centering\hspace{-2mm}
	\includegraphics[width=\linewidth]{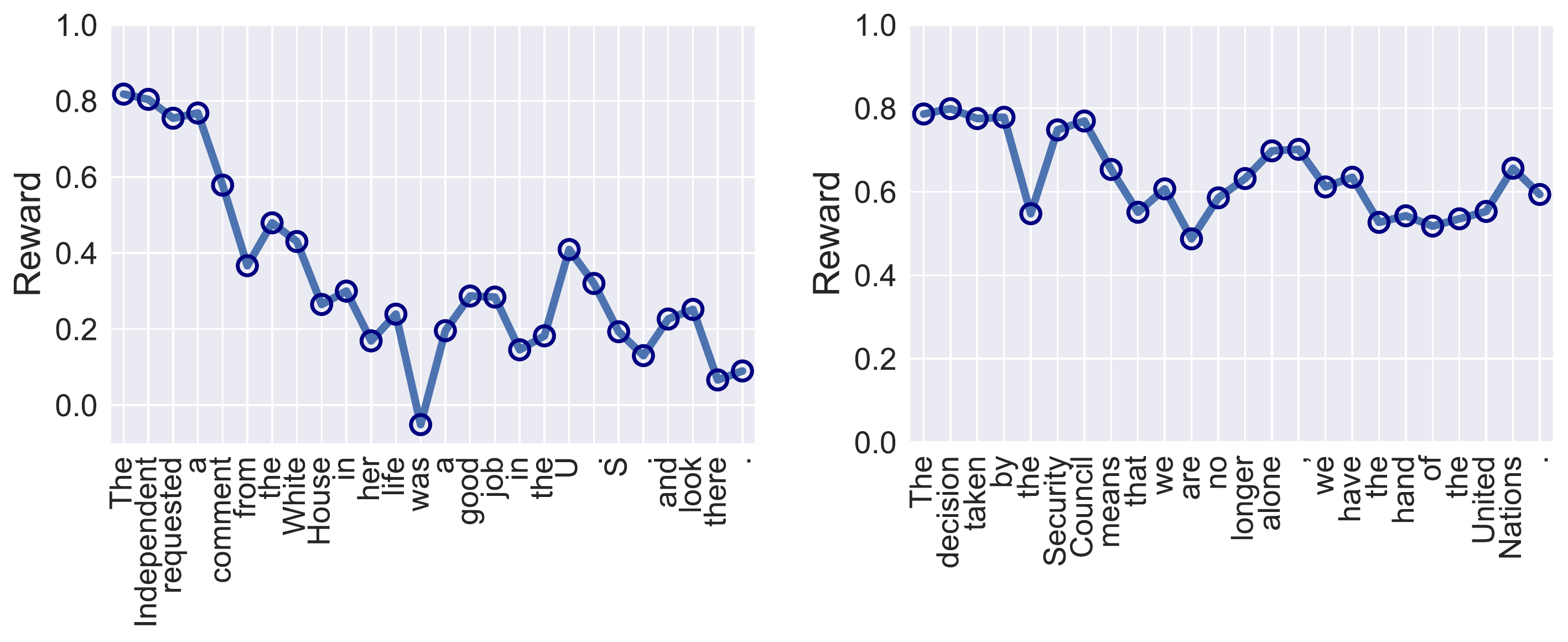} 
	\caption{Feature Matching Rewards Illustration.}\label{fig:rewards_ill}
\end{figure}

Figure \ref{fig:rewards_ill}(a) illustrates the feature-matching rewards during the generation, where it shows an example of failure generation at the initial RL-training stage, when two sentences are combined by the word `\emph{was}'. It is grammatically wrong to select `\emph{was}' for the generator, thus the guider network generates a negative rewards. We can see that the rewards becomes lower with more time steps, which is consistent with the exposure bias. Figure \ref{fig:rewards_ill}(b) shows a successful generation, where the rewards given by the guider are relatively high (usually larger than 0.5). These observations validate that: (i) exposure bias exists in MLE training. (ii) RL training with exploration can help reducing the effects of exposure bias. (iii) Our proposed feature-matching rewards can provide meaningful guidance to maintain sentence structure and fluency.

\paragraph{More Generated Samples of Text Generation}
Table \ref{tab: wmt_appendix} lists more generated samples on the proposed GMGAN and its baselines. From the experiments, we can see, (i) SeqGAN tends to generate shorter sentences, and the readability and fluency is very poor. (ii) LeakGAN tends to generate very long sentences, and usually longer than the original sentences. However, even with good locality fluency, its sentences usually are not semantically consistent. By contrast, our proposed GMGAN can generate sentences with similar length to the original sentences, and has good readability and fluency. This is also validated in the Human evaluation experiment.
\paragraph{Generated Samples of Style Transfer}
We show more examples of style transfer of our proposed methods in Table~13 and Table 14
, which contains 30 sentences of sentiment transfer. As can be seen, our proposed method can maintain most contents of the original sentences after the style transfer.

\section{Style Transfer with Guider Network}
Our framework naturally provides a way for style transfer, where the guider network plays the role of sentiment selection, and the generator only focus on generating meaningful sentence without considering the sentiments. 

To make the guider network focus on the guidance of sentiments, we use the label vector $l$ as the initial state $\sv^G_0$ of the guider network. Especially, at each step $t$, we feed the current sentence representation $\fv_t$ and label $l$ into the guider network:
\begin{align}
\hat{\fv}_{t+c} = G^{\psi}(\sv^G_{t-1}, [\fv_{t}, l])\,.
\end{align}

For the generator, we put an adversarial regularizer on the encoded latent $\sv_0(X)$ and penalize it if it contains the sentiment information. 
Intuitively, the generator gives candidate words represented by $O_t$, while the guider based on the sentiment information to make choice implicitly by $w_t$. So the sentiment information are contained in $w_t$, while the contents of the original sentences are represented by $O_t$. 

\section{Comments on the Guider network}
\subsection{Final Rewards of Unconditional Sequence Generation}
One can adopt many GAN variants to define final rewards. However, it is typically computational expensive and difficult to generalize. For example, it is extremely difficult to train a single discriminator to discriminate the partial generated and real sentences in the generative process. The CNN extracts features of the current sentence, which is an abstraction in the feature space. The LSTM guider network then takes this sequence of features to produce predictions.
In unconditional generation, the evaluation metric such as BLEU is not designed to provide reasonable guidance for the generator, since the references are all real sentences. Besides, even with a positive or negative feedback, every word generation is not consistently positive or negative.
%
\subsection{Connections between RL and Sequence Generation}
\paragraph{Exposure bias and exploration}
Reinforcement learning techniques are well known to be suitable for maximizing non-differentiable evaluation metrics, and thus can be applied for sequence-level optimization to solve the {\em exposure bias} problem \cite{ranzato2015sequence}. Exposure bias is avoided because the RL training is done by feeding the generated words back as input to the next time step. Sequence generation with RL also benefits from the exploration process, because it endows the model with the ability to explore unknown states in the generation process, encouraging the algorithm to generate more-diverse sentences. In this sense, stochastic policy could be helpful, since it better balances the exploration-exploitation trade-off.
\paragraph{RL provides long-term rewards}
When using an additional Q-function or advantage function constructed from immediate rewards, the generator is trained to focus on achieving higher long-term rewards. 
\paragraph{MLE training as Imitation learning}
MLE learning can be regarded as a naive way of imitation learning. The main issue of supervised imitation learning approach is that it fails to generalize to unseen situations and cannot learn to recover from failures, similar to the exposure bias in sequence generation \cite{ranzato2015sequence}. 
\paragraph{Discriminator as Inverse RL}
A discriminator in the text generation framework is used to provide rewards for generated sentences. It can be regarded as an evaluator for the generator to give higher rewards for high quality generated sequences, and lower rewards otherwise. This is similar to inverse RL, which try to learn a model for the rewards.
\subsection{Guider Network and Model-based RL}
Guider network can be regarded as a model of the sequence-generation environments, namely  
the model of dynamics. It takes current $\sv_t$ and $\av_t$ as input, and outputing an estimate of the next state $\sv_{t+\triangle t}$ at time $t+\triangle t$. In the sequence generation setting, when $\triangle t=1$, we can exactly get the feature representation of the current generated sentence if the guider does not help the word selection. If not, we cannot exactly get this feature extraction since the guider's prediction partly determine next token.
In practice, we use $\triangle t=c=4$, to give the guider planning ability, to help for word selection and guide sentence generation.

\section{Algorithum Overview}
\begin{algorithm}
	\caption{Guider Matching Generative Adversarial Network (GMGAN)}
	\label{algo:algo1}
	\begin{algorithmic}[1]
		\REQUIRE generator policy $\pi^{\phi}$; discriminator $D_\theta$; guider network $G^\psi$; a sequence dataset $\mathcal{S}=\{X_{1 \ldots T}\}$.
		\STATE Initialize $G^\psi$, $\pi^{\phi}$, $D^\theta$ with random weights.
		\STATE Pretrain generator $\pi^{\phi}$, guider $G^\psi$ and discriminator $D^\theta$ with MLE loss.
		\REPEAT 
		\FOR {g-steps}
		\STATE{Generate a sequence $Y_{1\ldots T} \sim \pi^{\phi}$.}
		\STATE {Compute $Q_t$ via (5), and update $\pi^\phi$ with policy gradient via (8).}
		\ENDFOR
		\FOR {d-steps}
		\STATE {Generate a sequences from $\pi^{\phi}$.}
		\STATE {Train discriminator $D_\theta$.}
		\ENDFOR
		\UNTIL{GMGAN converges}
	\end{algorithmic}
\end{algorithm}
\begin{algorithm}
	\caption{Guider Matching Sequence Training (GMST)}
	\label{algo:algo1}
	\begin{algorithmic}[1]
		\REQUIRE generator policy $\pi^{\phi}$; discriminator $D_\theta$; guider network $G^\psi$; a sequence dataset $\mathcal{S}=\{Y_{1 \ldots T}\}$ and its condition information $\mathcal{I}=\{X\}$
		\STATE Initialize $G^\psi$, $\pi^{\phi}$, $D^\theta$ with random weights.
		\STATE Pretrain generator $\pi^{\phi}$, guider $G^\psi$ and discriminator $D^\theta$ with MLE loss.
		\REPEAT 
		\STATE{Generate a sequence $Y_{1\ldots T} \sim \pi^{\phi}$.}
		\STATE{Compute evaluation scores based on references.}
		\STATE {Compute $Q_t^s$ via (6), and update $\pi^\phi$ with policy gradient via (8).}
		\UNTIL{GMST converges}
	\end{algorithmic}
\end{algorithm}

\newpage

\begin{table*}[h]
	\centering
	\begin{adjustbox}{scale=0.6,tabular=l | p{20.4cm},center}
		\textbf{Method} & \textbf{Generated Examples}\\
		\hline
		Real Data &
		What this group does is to take down various different websites it believes to be criminal and leading to terrorist acts .\newline
		Over 1 , 600 a day have reached Greece this month , a higher rate than last July when the crisis was already in full swing .\newline
		" We ' re working through a legacy period , with legacy products that are 10 or 20 years old ," he says .\newline
		' The first time anyone says you need help , I ' m on the defensive , but that ' s all that I know .\newline
		Out of those who came last year , 69 per cent were men , 18 per cent were children and just 13 per cent were women .\newline
		He has not played for Tottenham ' s first team since and it is now nearly two years since he completed a full Premier League match for the club .\newline
		So you have this man who seems to represent this way to live and how to be a good citizen of the world .\newline
		CNN : You made that promise , but it wasn ' t until 45 years later that you acted on it .\newline
		This is a part of the population that is notorious for its lack of interest in actually showing up when the political process takes place .\newline
		They picked him off three times and kept him out of the end zone in a 22 - 6 victory at Arizona in 2013 .\newline
		The treatment was going to cost £ 12 , 000 , but it was worth it for the chance to be a mum .\newline
		But if black political power is so important , why hasn ' t it made more of a difference in the lives of poor black people in Baltimore such as Gray ?\newline
		Local media reported the group were not looking to hurt anybody , but they would not rule out violence if police tried to remove them .\newline
		The idea was that couples got six months ' leave per child with each parent entitled to half the days each .\newline
		The 55 to 43 vote was largely split down party lines and fell short of the 60 votes needed for the bill to advance .\newline
		Taiwan ' s Defence Ministry said it was " aware of the information ," and declined further immediate comment , Reuters reported .\newline
		I ' m racing against a guy who I lost a medal to - but am I ever going to get that medal back ?\newline
		Others pushed back their trips , meaning flights early this week are likely to be even more packed than usual .\newline
		" In theory there ' s a lot to like ," Clinton said , " but ' in theory ' isn ' t enough .\newline
		If he makes it to the next election he ' ll lose , but the other three would have lost just as much .
		\\
		\hline
		SeqGAN & Following the few other research and asked for " based on the store to protect older , nor this . \newline
		But there , nor believe that it has reached a the person to know what never - he needed . \newline
		The trump administration later felt the alarm was a their doctors are given . \newline
		We have been the time of single things what people do not need to get careful with too hurt after wells then . \newline
		If he was waited same out the group of fewer friends a more injured work under it . \newline
		It will access like the going on an " go back there and believe . \newline
		Premier as well as color looking to put back on a his is . \newline
		So , even though : " don ' t want to understand it at an opportunity for our work . \newline
		I was shocked , nor don ' t know if mate , don ' t have survived ,  \newline
		So one point like ten years old , but a sure , nor with myself more people substantial . \newline
		And if an way of shoes of crimes the processes need to run the billionaire . \newline
		Now that their people had trained and people the children live an actor , nor what trump had . \newline
		However , heavily she been told at about four during an innocent person . \\
		\hline
		LeakGAN & The country has a reputation for cheap medical costs and high - attack on a oil for more than to higher its - wage increase to increase access to the UK the UK women from the UK ' s third nuclear in the last couple of weeks .\newline
		I ' ve been watching it through , and when the most important time it is going to be so important .\newline
		I ' m hopeful that as that process moves along , that the U . S . Attorney will share as much as far as possible .\newline
		The main thing for should go in with the new contract , so the rest of the Premier League is there to grow up and be there ," she said .\newline
		I think the main reason for their sudden is however , I didn ' t get any big thing ," he says , who is the whole problem on the U . S . Supreme Court and rule had any broken .\newline
		The average age of Saudi citizens is still very potential for the next year in the past year , over the last year he realised he has had his massive and family and home .\newline
		" I think Ted is under a lot of people really want a " and then the opportunity to put on life for security for them to try and keep up .\newline
		The new website , set to launch March 1 , but the U . S is to give up the time the case can lead to a more than three months of three months to be new home .\newline
		It ' s a pub ; though it was going to be that , but , not , but I am not the right thing to live ," she said .\newline
		" I ' m not saying method writing is the only way to get in the bedroom to get through the season and we ' ll be over again ," he says .\newline
		I ' m not suggesting that our jobs or our love our years because I have a couple of games where I want it to be .\newline
		The German government said 31 suspects were briefly detained for questioning after the New Year ' s Eve trouble , among them not allowed to stay in the long - term .\newline
		It was a punishment carried out by experts in violence , and it was hard to me he loved the man and he ' s got off to support me in the future .\newline
		" I ' ve known him , all that just over the last two weeks and for the last 10 years , I ' ll have one day of my life ," she said .\newline
		The main idea behind my health and I think we saw in work of our country was in big fourth - up come up with a little you ' ve ever .\newline
		he Kings had needed scoring from the left side , too , and King has provided that since his return are the of the first three quarters of the game .\newline
		The average number of monthly passengers arriving at the University of January 1 . 1 million people and another average visit men were on the year .\newline
		It ' s going to be a good test for us and we are on the right way to be able to get through it on every day on the year .\\
		\hline
		GMGAN & " I ' m actually going to take my baby shopping and get him go back and never let them go into the property .\newline
		The ban will continue in several European countries that were occupied by Nazi Germany , including Austria and the Netherlands .\newline
		But 2015 was also a year when people again took to the streets to protest corruption - people across the globe sent a strong signal to those in power : it is .\newline
		But the best advice , especially if you ' re starting out , you can get a feel that works for them sure .\newline
		She said : " To those for those who were in music , time you never say to live here .\newline
		We ' re certainly going to have to prepare and coach the team a lot better than we did that night .\newline
		But the benefits from the UK - are being prepared to have used in Australian middle for alcohol consumption and six weeks or less than the year in easier times .\newline
		It ' s a very well - I ' ve really had and it makes me feel like a lot of beach for what I ' m doing ," she says .\newline
		We ' re creating the space for them to think about what their choices are , because at the end of the day , the players will be OK .\newline
		He said E . On customers could wait for the small reduction in their bill or shop around and save more than £ 300 a year .\newline
		" I ' m actually going to take my baby shopping and get him go back and never let them go into the property .\newline
		' I was actually surprised that he got some other as a big that I would be as a big storm ," she said .\newline
		He returned to work on his father ' s maintenance crews - while his head is about to the Democratic nomination , a 15 - year - old who did not learn as a way to everything play behind responsible for the guns , they sell with us later .\newline
		But the eye of the storm was China , where the main index in Shanghai lost 19 \% of its value in the same period .\newline
		The capital could get 15 - 20 inches , Philadelphia could see 12 to 18 and New York City and Long Island could get 8 to 10 .\newline
		A Virginia couple was surprised after receiving a letter their son sent almost 11 years ago while serving in Iraq .\newline
		I ' m looking forward to going to battle with those guys all year long and for the rest of our careers .
		A new ISIS video experience just - aged after the three - year period , and two of three inmates - are thought strongly , the U . N ? and campaign .\newline
		" I ' m well aware that little ones can get into trouble and well ," Ms Turner ' s business .\newline
		It ' s a wonder the producers of this year ' s show did not sign up someone from the world of sport .\\
		\hline
		
	\end{adjustbox}
	\caption{
		Generated Examples on EMNLP2017 WMT.
	}
	\label{tab: wmt_appendix}
\end{table*}
\newpage

\begin{table*}[h]
	\centering
	\begin{adjustbox}{scale=0.9,tabular=l  p{13.4cm},center}
				\toprule[1.2pt]
				Original: & i 'm so lucky to have found this place ! \\
				Transferred: &	i 'm so embarrassed that i picked this place . \\
				\hline
				Original: & awesome place , very friendly staff and the food is great !  \\
				Transferred: & disgusting place , horrible staff and extremely rude customer service . \\
				\hline
				Original: & this was my first time trying thai food and the waitress was amazing ! \\
				Transferred: & this was my first experience with the restaurant and we were absolutely disappointed .  \\
				\hline
				Original: & thanks to this place ! \\
				Transferred: & sorry but this place is horrible . \\
				\hline
				Original: & the staff was warm and friendly . \\
				Transferred: & the staff was slow and rude .  \\
				\hline
				Original: & great place and huge store . \\
				Transferred: & horrible place like ass screw .  \\
				\hline
				Original: & the service is friendly and quick especially if you sit in the bar . \\
				Transferred: &	the customer service is like ok - definitely a reason for never go back .. \\
				\hline
				Original: & everything is always delicious and the staff is wonderful . \\
				Transferred: & everything is always awful and their service is amazing . \\
				\hline
				Original: & best place to have lunch and or dinner . \\
				Transferred: & worst place i have ever eaten . \\
				\hline
				Original: & best restaurant in the world ! \\
				Transferred: & worst dining experience ever !  \\
				\hline
				Original: & you 'll be back ! \\
				Transferred: &	you 're very disappointed ! \\
				\hline
				Original: & you will be well cared for here ! \\
				Transferred: & you will not be back to spend your money .  \\
				\hline
				Original: & they were delicious ! \\
				Transferred: & they were overcooked .  \\
				\hline
				Original: & seriously the best service i 've ever had . \\
				Transferred: & seriously the worst service i 've ever experienced . \\
				\hline
				Original: & it 's delicious ! \\
				Transferred: &	it 's awful . \\
				\bottomrule[1.2pt] 
	\end{adjustbox}
	\label{tab: trans_samples1}
	\caption{Sentiment transfer samples on Yelp dataset (positive $\to$ negative).}
\end{table*}

\begin{table*}[h]
	\centering
	\begin{adjustbox}{scale=0.9,tabular=l  p{13.4cm},center}
		\toprule[1.2pt]
		Original: & gross !  \\
		Transferred: & amazing !  \\
		\hline
		Original: & the place is worn out . \\
		Transferred: &	the place is wonderful . \\
		\hline
		Original: & very bland taste .  \\
		Transferred: & very fresh . \\
		\hline
		Original: & terrible service ! \\
		Transferred: & great customer service !  \\
		\hline
		Original: & this place totally sucks . \\
		Transferred: & this place is phenomenal .  \\
		\hline
		Original: & this was bad experience from the start . \\
		Transferred: & the food here was amazing good .  \\
		\hline
		Original: & very rude lady for testing my integrity . \\
		Transferred: & very nice atmosphere for an amazing lunch ! \\
		\hline
		Original: & they recently renovated rooms but should have renovated management and staff . \\
		Transferred: & great management and the staff is friendly and helpful .  \\
		\hline
		Original: & this store is not a good example of sprint customer service though . \\
		Transferred: &	this store is always good , consistent and they 're friendly . \\
		\hline
		Original: & one of my least favorite ross locations . \\
		Transferred: & one of my favorite spots .  \\
		\hline
		Original: & horrible in attentive staff . \\
		Transferred: & great front desk staff !  \\
		\hline
		Original: & the dining area looked like a hotel meeting room . \\
		Transferred: & the dining area is nice and cool .  \\
		\hline
		Original: & never ever try to sell your car at co part ! \\
		Transferred: & highly recommend to everyone and recommend this spot for me ! \\
		\hline
		Original: & i ordered the filet mignon and it was not impressive at all . \\
		Transferred: &	i had the lamb and it was so good .\\
		\bottomrule[1.2pt] 
	\end{adjustbox}
	\label{tab: trans_samples}
	\caption{Sentiment transfer samples on Yelp dataset (negative $\to$ positive).}
\end{table*}


\end{document}